\documentclass[runningheads]{llncs}
\usepackage{cite}
\usepackage{url}

\usepackage{graphicx}
\usepackage{algorithm}
\usepackage{algorithmic}
\usepackage{newfloat}
\usepackage{listings}
\usepackage{cite}
\usepackage{multirow}
\usepackage{multicol}
\usepackage{amsmath,amssymb,amsfonts}
\usepackage[figuresright]{rotating}
\usepackage{subfigure}
\usepackage{color}
\usepackage{hyperref}
\begin{document}
	\title{DADFNet: Dual Attention and Dual Frequency-Guided Dehazing Network for Video-Empowered Intelligent Transportation}
	%
	%
	\author{
		Yu Guo\textsuperscript{\rm 1},
		Ryan Wen Liu\textsuperscript{\rm 1},
		Jiangtian Nie\textsuperscript{\rm 2}, 
		Lingjuan Lyu\textsuperscript{\rm 3}, 
		Zehui Xiong\textsuperscript{\rm 4}, 
		Jiawen Kang\textsuperscript{\rm 5}, 
		Han Yu\textsuperscript{\rm 2},
		Dusit Niyato\textsuperscript{\rm 2}
	}
	\authorrunning{Y. Guo et al.}
	%
	\institute{School of Navigation, Wuhan University of Technology, Wuhan, China \and
		School of Computer Science and Engineering, Nanyang Technological University, Singapore \and Sony AI, Tokyo, Japan \and Information Systems Technology and Design Pillar, Singapore University of Technology and Design, Singapore \and School of Automation, Guangdong University of Technology, Guangzhou, China
		\email{\{yuguo, wenliu\}@whut.edu.cn, jnie001@e.ntu.edu.sg, lingjuanlvsmile@gmail.com, zehui\_xiong@sutd.edu.sg, kavinkang@gdut.edu.cn, \{han.yu, dniyato\}@ntu.edu.sgger.com}}
	\maketitle              
	\begin{abstract}
		Visual surveillance technology is an indispensable functional component of advanced traffic management systems. It has been applied to perform traffic supervision tasks, such as object detection, tracking and recognition. However, adverse weather conditions, e.g., fog, haze and mist, pose severe challenges for video-based transportation surveillance. To eliminate the influences of adverse weather conditions, we propose a dual attention and dual frequency-guided dehazing network (termed DADFNet) for real-time visibility enhancement. It consists of a dual attention module (DAM) and a high-low frequency-guided sub-net (HLFN) to jointly consider the attention and frequency mapping to guide haze-free scene reconstruction. Extensive experiments on both synthetic and real-world images demonstrate the superiority of DADFNet over state-of-the-art methods in terms of visibility enhancement and improvement in detection accuracy. Furthermore, DADFNet only takes $6.3$ ms to process a $1,920 \times 1,080$ image on the $2080$ Ti GPU, making it highly efficient for deployment in intelligent transportation systems.
		
		\keywords{Visual surveillance \and Deep learning \and Intelligent transportation system \and Image dehazing \and Attention mechanism.}
	\end{abstract}
	\section{Introduction}
\begin{figure}[t]
		\centering
		\includegraphics[width=1\linewidth]{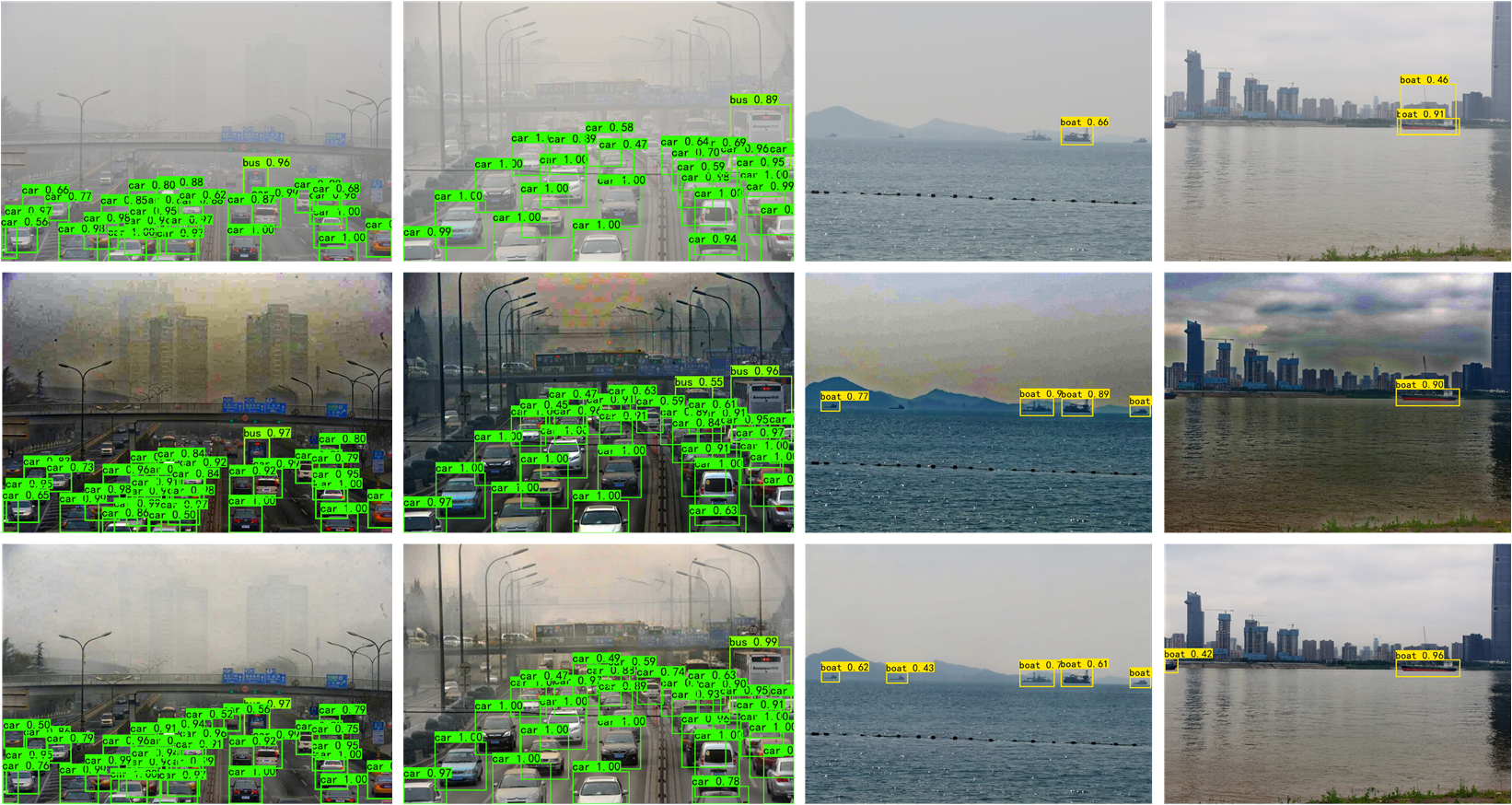}
		\caption{The detection results on original hazy images (Top) and restored images by  (Middle) DCP \cite{he2010single} and (Bottom) our DADFNet. Note that we select the YOLOv4 \cite{bochkovskiy2020yolov4} as the detection method, which is trained on the VOC.}
		\label{fig:detetion}
	\end{figure}

	As an indispensable component of information collection, visual sensors play a pivotal role in intelligent transportation systems (ITSs). Many technologies based on rich semantic information in visual data have been proposed to promote ITS development (e.g., object detection, recognition, tracking, and autonomous navigation, etc.). However, the videos/images captured outdoors are negatively affected by haze. This seriously limits the application of visual surveillance technologies. Fig. \ref{fig:detetion} provides four examples of detection failure in the road and maritime traffic scenarios under hazy conditions. Specifically, the signal captured by the observers is attenuated under hazy conditions due to the scattering and absorption of ambient light by turbid medium in the atmosphere. This degradation phenomenon reduces the contrast and color fidelity of the entire image. Meanwhile, the foreground and background become blurred and deformed, resulting in loss of the critical target information. As the distance between the observer and the target increases, the degradation phenomenon becomes more severe, which increases the risk of detection failures. For example, some distant ships in the third scene of Fig. \ref{fig:detetion} are disturbed by haze. As a result, these objects are not robustly detected. To deploy computer vision technology in ITS under hazy conditions for effective intelligent surveillance, it is necessary to design a practical image dehazing method.

	Though image dehazing has been studied for many years, developing a practical dehazer is challenging. Model-based dehazing methods usually recover images by estimating the parameters in the atmospheric scattering model. Therefore, the dehazing effect of these methods depends on whether the two parameters are accurately estimated simultaneously. Undoubtedly, it is ill-posed and challenging to estimate two terms from a single hazy image. Once the prior theory fails, the visibility of the recovered image will be unchanged or even worse. Fig. \ref{fig:detetion} shows four dehazing results obtained by the dark channel prior (DCP) method \cite{he2010single}. It can be clearly found that DCP easily causes color distortion of the water surface and sky region. 

	With the rapid evolution of graphics processing units (GPUs), data-driven methods have been widely used in low-level image processing tasks, e.g., denoising and low-light enhancement. Although data-driven dehazing strategies have received continuous attention, few methods are designed for traffic scenarios. Different from plain image processing tasks, the dehazer deployed in ITS should sufficiently restore the crucial information in the real-world traffic hazy environment on the premise of ensuring real-time operations.

	Based on the above analysis, we design a dual attention and dual frequency-guided dehazing network (termed DADFNet). In particular, DADFNet is mainly composed of two parts, i.e., dual attention module (DAM) for obtaining hazy distribution in color and pixel spaces and high-low frequency-guided sub-net (HLFN) for reconstructing clear scenes. 
	The main contributions of our proposed DADFNet are as follows:
	%
	\begin{itemize}
		\item We propose an efficient dehazer to improve the video/image visibility and the precision of high-level vision technology deployed in ITS under hazy conditions.
		\item DAM consists of a color attention (CA) and a multi-scale pixel attention (MSPA) to get the hazy distribution. 
		\item HLFN includes a high-low frequency generator to focus on the low-frequency structures and high-frequency details simultaneously.
		\item Numerous experiments have been performed to demonstrate the efficacy of the proposed dehazing method.
	\end{itemize}

	\section{Related Work}
	Single image dehazing is a hot topic in the field of image processing and many dehazing strategies have been proposed. This section will briefly provide a review of classic and current researches closely related to our works.
	\subsection{Model-Based Methods}
	Driven by physical imaging model \cite{narasimhan2000chromatic}, the model-based methods generally design manual priors based on certain empirical observations to estimate the haze-free image. To suppress haze, \cite{tan2008visibility} maximized the local contrast by developing a cost function in the framework of Markov random fields. However, the excessive enhancement of the contrast will cause scene distortion in the real-world dehazing task. To eliminate this negative influence, \cite{he2010single} proposed a dark channel prior (DCP) to reconstruct haze-free scenarios by optimizing the transmission. Unfortunately, DCP easily provides poor visual effects in certain situations, such as white scenes and sky regions. Therefore, many DCP-based dehazing methods \cite{zhu2018haze, kim2019fast, shu2019variational} are proposed for solving this issue. Furthermore, several prior-based strategies have been constructed, e.g., color-lines prior \cite{fattal2014dehazing}, color attenuation prior \cite{zhu2015fast}, non-local prior \cite{berman2016non}, etc. Undoubtedly, the performance of model-based dehazing methods strongly depends on the accuracy of prior knowledge, which will cause poor robustness in the complex real-world hazy scene. 
	%
	%
	\subsection{Data-Driven Methods}
	Unlike model-based methods, data-driven methods generally learn the hazy features from the training dataset by the deep neural network. In current literature, data-driven methods can be categorized into the parameter estimation method and the end-to-end method.

	On the basis of the atmospheric scattering model, the parameter estimation method reconstructs the clear scene by generating the transmission and atmospheric light. Early attempts, such as DehazeNet \cite{cai2016dehazenet} and MSANN \cite{ren2020single}, estimated the transmission map by convolutional neural network (CNN). However, atmospheric light is a non-adjustable parameter in these methods, which will seriously affect the dehazing effect. Therefore, \cite{li2017aod} designed an AODNet for estimating a novel parameter from the transformed atmospheric scattering model. Meanwhile, DCPDN \cite{zhang2018densely}, a densely connected pyramid dehazing network, is proposed to recovery haze-free images by generating the transmission and atmospheric light jointly. Although the performance of these methods has been proven, the simultaneous estimation of several parameters easily produces reconstruction loss, which magnifies the anomalies. Furthermore, it is challenging to obtain accurate ground truth of transmission and atmospheric light in real-world scenes.

	To reduce the reconstruction error and avoid the influence generated by estimation model parameters, many end-to-end CNN methods have recently been proposed to directly learn the mapping of hazy-to-clean images. For instance, \cite{ren2018gated} proposed a GFN model based on the fusion strategy. In particular, GFN can comprehensively consider the results of white balance (WB), contrast enhancement (CE), and gamma correction (GC) and fuse the results of all methods to generate the clean image by an encoder-decoder network. In 2020, \cite{qin2020ffa} designed a feature fusion attention network (FFANet), which consists of two attention modules for dealing with feature information from channel and pixel spaces, respectively. For the sake of better dehazing performance, \cite{qin2020ffa} constructed a contrastive learning-driven autoencoder-like framework (named AECRNet) by exploiting the negative information. Although these methods have satisfactory effects with varying degrees, it is tricky to deploy these methods in outdoor traffic scenarios. The difficulty of this issue is how to adequately eliminate the haze of particular scenes (e.g., road and maritime) in the premise of real-time performance.
	%
	%
	%
	%
	
	%
	%
	%
	%
	%
	\begin{figure*}[t]
		\centering
		\includegraphics[width=1\linewidth]{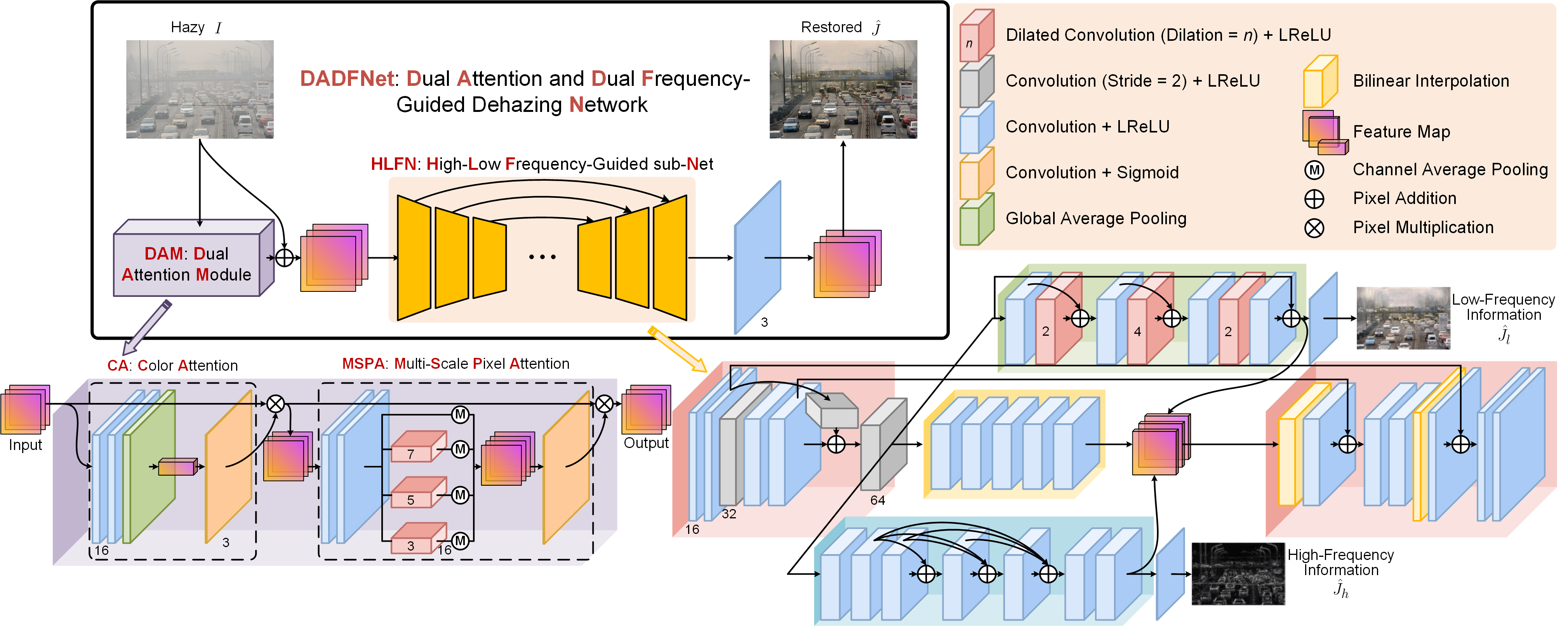}
		\caption{The architecture of our proposed dual attention and dual frequency-guided dehazing network (DADFNet). The DADFNet mainly consists of two parts, i.e., dual attention module (DAM) and high-low frequency-guided sub-net (HLFN). Note that LReLU denotes the leaky rectified linear unit function.}
		\label{fig:framework}
	\end{figure*}
	\section{DADFNet: Haze Visibility Enhancement Network}
	This section introduces the details of the proposed dehazing network. We refer to this network as dual attention and dual frequency-guided dehazing network (DADFNet). The framework of our proposed DADFNet is shown in Fig. \ref{fig:framework}. In particular, this network mainly consists of two parts, named dual attention module (DAM) and high-low frequency-guided sub-net (HLFN).
	\subsection{Dual Attention Module}
	Plain CNN treats each color (i.e., R, G, and B) and pixel spaces with the same attention. However, the hazy distribution in the image domain is often non-homogeneous or associated with field depth. To focus on the hazy features in the color and pixel, we design a dual attention module (DAM), including a color attention (CA) and a multi-scale pixel attention (MSPA). 
	\subsubsection{Color Attention (CA)}
	Unlike channel attention methods \cite{qin2020ffa, woo2018cbam}, our proposed CA considers that R, G, and B color spaces have different attention in image dehazing. As shown in Fig. \ref{fig:framework}, the 16-channel feature map $\bar{F}_{c}$ is first generated by two convolution layers $Conv$ from the input hazy image $I$, and the leaky rectified linear unit (LReLU) activation function $\mathcal{F}_{l}$ is deployed after each $Conv$, i.e.,
	\begin{equation}\label{Eq:CA1}
	\bar{F}_{c} = \mathcal{F}_{l}(Conv(\mathcal{F}_{l}(Conv(I)))).
	\end{equation} 
	Secondly, we adopt global average pooling to get the weighted information $\tilde{F}_{c}$ of all channels, which can be expressed as
	\begin{equation}\label{Eq:CA2}
	\tilde{F}^{k}_{c} = \frac{1}{N}\sum_{i=1}^{N}\bar{F}^{k}_{c}(i),
	\end{equation} 
	where $\tilde{F}^{k}_{c}$ is the feature map of $k$-th channel in the $\tilde{F}_{c}$, $N$ denotes the number of pixels in $\bar{F}_{c}$. Finally, the weight tensor $F_{c}$ of R, G, and B color channels are obtained by a $Conv$ and Sigmoid function $\mathcal{F}_{s}$, which can be given by
	\begin{equation}\label{Eq:CA3}
	F_{c} = \mathcal{F}_{s}(Conv(\tilde{F}_{c})).
	\end{equation}
	\subsubsection{Multi-Scale Pixel Attention (MSPA)}
	Since the hazy distribution is generally non-homogeneous in the pixel space, we design a multi-scale pixel attention (MSPA) module. In particular, MSPA takes the hazy image weighted by CA as input. The weighted hazy image $I^{\ast}$ can be written as follows
	\begin{equation}\label{Eq:MSPA1}
	I^{\ast} = I \otimes F_{c},
	\end{equation}
	with $\otimes$ being the operator of pixel-wise multiplication.
	
	Similar to the CA, the proposed MSPA first adopts two convolution layers with LReLU to generate the 16-channel feature map $\bar{F}_{p}$, i.e.,
	\begin{equation}\label{Eq:MSPA2}
	\bar{F}_{p} = \mathcal{F}_{l}(Conv(\mathcal{F}_{l}(Conv(I^{\ast})))).
	\end{equation} 

	To fully extract the hazy information in different receptive fields, we construct an inception-based feature extractor. In particular, three dilated convolutions $DConv_{m}$ with rate $m \in \{3, 5, 7\}$ are respectively deployed to extract the multi-scale features of $\bar{F}_{p}$, which can be defined as
	\begin{equation}\label{Eq:MSPA3}
	g_{m} = \mathcal{F}_{l}(DConv_{m}(\bar{F}_{p})),
	\end{equation} 
	where $g_{m}$ is the output of $DConv_{m}$. Then, three feature maps and $\bar{F}_{p}$ are merged into a $4$-channel feature map, i.e.,
	\begin{equation}\label{Eq:MSPA4}
	\tilde{F}_{p} = [\omega(\bar{F}_{p}),\omega(g_{3}),\omega(g_{5}),\omega(g_{7})],
	\end{equation}
	with $\omega$ being the operator for calculating the mean value of channel-wise, termed channel average pooling. The single-channel weight tensor $F_{p}$ of all pixels are generated by a $Conv$ which regards Sigmoid $\mathcal{F}_{s}$ as an activation function, i.e.,

	\begin{equation}\label{Eq:MSPA5}
	F_{p} = \mathcal{F}_{s}(Conv(\tilde{F}_{p})),
	\end{equation}
	Mathematically, the output image $I^{\star}$ weighted by MSPA is expressed as follows
	\begin{equation}\label{Eq:MSPA6}
	I^{\star} = I^{\ast} \otimes F_{p}.
	\end{equation}
	\subsection{High-Low Frequency-Guided Sub-Net}
	Although many CNN-enabled dehazing methods have been developed, the restored images still suffer from detail loss. Therefore, we construct a high-low frequency-guided sub-net (HLFN) for better detail and structure extraction performance. As shown in Fig. \ref{fig:framework}, the high-low frequent-guided sub-net is an encoder-decoder network. The baseline of HLFN is a typical three-scale U-Net \cite{ronneberger2015u}, which is composed of convolution and LReLU. In particular, we adopt feature reuse strategies in the encoder to obtain high-level and low-level features. Meanwhile, we embed a high-low frequency generator in HLFN for stronger image reconstruction ability.
	\subsubsection{High-Low Frequency Generator} 
	To recover the high-frequency details and low-frequency structure concurrently, we design a high-low frequency generator to guide the reconstruction of sharp images. The proposed generator contains high-frequency and low-frequency networks, each of which includes $7$ convolutional layers. Due to the inconsistent functions, the high-frequency and low-frequency networks use different architectures. Specifically, the low-frequency network is a dense residual network to avoid the loss of high-frequency information (e.g., texture and details) through feature reuse. For example, let $F_2$, $F_3$, $F_4$, and $F_5$ be the output of the $2$-nd, $3$-rd, $4$-th, and $5$-th convolutional layers in the low-frequency network, respectively, the output of the $6$-th convolutional layer $F_6$ can be obtained by
	\begin{equation}\label{Eq:HLN}
	F_{6} = \mathcal{F}_{l}(Conv(F_{2}+F_{3}+F_{4}+F_{5})),
	\end{equation}
	with $\mathcal{F}_{l}$ and $Conv$ being the LReLU activation function and convolution, respectively. To enable the proposed DADFNet recover more natural color, contrast, and structure, the low-frequency network is designed. In particular, the low-frequency network is a simplified residual network, which contains two local residuals and a global residual. Moreover, this network contains three dilated convolutions to generate multi-scale structural information. Finally, the two feature maps calculated by the high-low frequency generator are, respectively, processed by convolution to obtain high-frequency and low-frequency RGB images.
	\subsection{Loss Function}
	Undoubtedly, it is essential to design an ideal loss function for seeking better reconstruction effect. Therefore, we construct a hybrid loss function for concurrently supervising the outputs of HLFN $\hat{J}$ and high-low frequency generator $(\hat{J}_{h}, \hat{J}_{l})$. In the image restoration task, the smooth L1 loss is less sensitive to outliers than L2 and has a faster convergence than L1. Therefore, we first use $\mathcal{L}_{sl1} = L_{sl1}(J-\hat{J})$ to constrain the ground truth $J$ and final output $\hat{J}$, where $L_{sl1}$ means the smooth L1 operator. Moreover, the outputs of high-low frequency generator $(\hat{J}_{h}, \hat{J}_{l})$ are supervised by $\mathcal{L}_{sl1}^{hl} = L_{sl1}(G_{h}(J)-\hat{J}_{h}) + L_{sl1}(G_{l}(J)-\hat{J}_{l})$, where $G_{h}$ and $G_{l}$ are Fourier transform-based high-pass filter and low-pass filter, respectively. However, the smooth L1 loss preserves the color and luminance of each region with equal weight. To improve the contrast of high-frequency regions in the image, we adopt $\mathcal{L}_\text{MS-SSIM} = 1 - \text{MS-SSIM}(J, \hat{J})$ for fully capturing the hazy characteristics, with $\text{MS-SSIM}$ being the Multi-scale structural similarity operation (MS-SSIM). Please refer to \cite{wang2017fast, wang2003multiscale} and references therein for more details on smooth L1 and MS-SSIM.

	To fully extract potential information from high-level feature space, we apply a perceptual loss based on the VGG-$16$ network \cite{simonyan2014very} to construct fine details. In particular, this VGG-$16$ network is pre-trained on the ImageNet. The perceptual loss is described as
	\begin{equation}\label{Eq:PL1}
	{\mathcal{L}_{per} = \frac{1}{3} \sum_{r} \frac{| \phi_{r}(J) - \phi_{r}(\hat{J}) | }{N_{k}},}
	\end{equation}
	where $\phi_{r}$ denotes the output of $r$-th layer in VGG-$16$. In this work, we set $r \in \{relu1\_2, relu2\_2, relu3\_3\}$. $N_{r}$ represents the output size of the $r$-th layer. Different from the traditional perceptual loss \cite{johnson2016perceptual}, we replace L2 loss with L1 loss for better dehazing effect.
	\begin{table*}[t]
		\setlength{\tabcolsep}{1.3pt}
		\tiny
		\centering
		\caption{Detailed configuration of the discriminator. Note that ``GAP'', ``$\mathcal{F}_{l}$'', ``$\mathcal{F}_{s}$'', and ``BN'' denote the global average pooling, leaky rectified linear unit (LReLU) function, sigmoid function, and batch normalization, respectively.}
		\begin{tabular}{ccccccccccccc}
			\hline
			& $Conv$1     & $Conv$2 & $Conv$3 & $Conv$4 & $Conv$5 & $Conv$6 & $Conv$7 & $Conv$8 & $Conv$9       & GAP & $Conv$10   & $Conv$11 \\ \hline \hline
			Input      & $J(\hat{J})$ & $Conv$1 & $Conv$2 & $Conv$3 & $Conv$4 & $Conv$5 & $Conv$5 & $Conv$5 & $Conv$5/6/7/8 & $Conv$9   & GAP & $Conv$10 \\
			Channel       & 16 & 16 & 32 & 32 & 64 & 64 & 64 & 64 & 64 & 64 & 64 & 64 \\
			Size          & 3  & 3  & 3  & 3  & 3  & 3  & 3  & 3  & 3  & --   & 1  & 1  \\
			Stride        & 1  & 2    & 1  & 2    & 1  & 1  & 1  & 1  & 1  & --   & 1  & 1  \\
			Padding       & 1  & 1  & 1  & 1  & 1  & 3  & 5  & 7  & 1  & --   & 0  & 0  \\
			Dilation      & 1  & 1  & 1  & 1  & 1  & 3  & 5  & 7  & 1  & --   & 1  & 1  \\
			Activation & $\mathcal{F}_{l}$        & $\mathcal{F}_{l}$    & $\mathcal{F}_{l}$    & $\mathcal{F}_{l}$    & $\mathcal{F}_{l}$    & $\mathcal{F}_{l}$    & $\mathcal{F}_{l}$    & $\mathcal{F}_{l}$    & $\mathcal{F}_{l}$          & --         & $\mathcal{F}_{l}$      & $\mathcal{F}_{s}$  \\
			Normalization & --   & BN   & BN   & BN   & BN   & BN   & BN   & BN   & BN   & --   & --   & --   \\ \hline
		\end{tabular} \label{table:gan}
	\end{table*}

	Furthermore, we add additional adversarial loss to generate more realistic results. The network architecture of the discriminator is shown in Table \ref{table:gan}. In particular, our proposed discriminator includes three dilated convolutions with different rates to extract the multi-scale features. Based on the constructed discriminator, we implemented an adversarial loss. Let the restored image $\hat{J}$ be input, the adversarial loss can be defined as 
		\begin{equation}\label{Eq:AL1}
		{\mathcal{L}_{adv} = -\frac{1}{W}\sum_{q=1}^{W} \log(D(\hat{J_{q}})).}
		\end{equation}
	Here, $D(\cdot)$ is the discriminator, $D(\hat{J_{q}})$ represents the confidence that the restored image $\hat{J_{q}}$ is the real haze-free image, $W$ denotes the quantity of the input image. As a result, the loss function of the proposed HLFN is summarized as follows
	\begin{equation}\label{Eq:TL1}
	{\mathcal{L}_\text{HLFN} = \lambda_{1} \mathcal{L}_{sl1} + \lambda_{2} \mathcal{L}_{sl1}^{hl} + \lambda_{3} \mathcal{L}_\text{MS-SSIM} + \lambda_{4} \mathcal{L}_{per} + \lambda_{5} \mathcal{L}_{adv}.}
	\end{equation}
	\section{Experimental Results}
	In this section, both synthetic and real-world dehazing experiments are carried out to verify the superior imaging effect of our method in outdoor traffic scenarios. We select $10$ state-of-the-art dehazing methods as the competitive methods, including DCP\cite{he2010single}, CAP \cite{zhu2015fast}, GRM \cite{chen2016robust}, HL \cite{berman2018single}, F-LDCP \cite{zhu2018haze}, DehazeNet \cite{cai2016dehazenet}, MSCNN \cite{ren2020single}, AODNet \cite{li2017aod}, GCANet \cite{chen2019gated}, and FFANet \cite{qin2020ffa}. Meanwhile, we implement object detection experiments after dehazing to illustrate that DADFNet can improve the precision of high-level vision tasks. Note that all parameters of these methods are provided by the authors' codes to guarantee fair comparison. Moreover, we adopt peak signal-to-noise ratio (PSNR), structural similarity index (SSIM), learned perceptual image patch similarity (LPIPS), and mean average precision (mAP) for evaluating the performance of various methods on different tasks.
	\begin{table*}[t]
		\setlength{\tabcolsep}{5pt}
		\centering
		\caption{PSNR, SSIM, and LPIPS results of various methods on the datasets of $50$ RESIDE and $50$ Seaships. The best results are in \textbf{bold}.}
		\begin{tabular}{cccc|ccc}
			\hline
			\multirow{2}{*}{Method} & \multicolumn{3}{c|}{RESIDE \cite{li2018benchmarking}}                          & \multicolumn{3}{c}{Seaships \cite{shao2018seaships}}                        \\
			& PSNR$\uparrow$ & SSIM$\uparrow$ & LPIPS$\downarrow$ & PSNR$\uparrow$ & SSIM$\uparrow$ & LPIPS$\downarrow$ \\ \hline \hline
			DCP \cite{he2010single}                    & 15.56          & 0.844          & 0.128             & 13.59          & 0.750          & 0.223             \\
			CAP \cite{zhu2015fast}                    & 20.55          & 0.914          & 0.061             & 21.08          & 0.929          & 0.070             \\
			GRM \cite{chen2016robust}                    & 18.44          & 0.857          & 0.144             & 17.65          & 0.845          & 0.207             \\
			HL \cite{berman2018single}                     & 19.57          & 0.894          & 0.088             & 18.38          & 0.855          & 0.156             \\
			F-LDCP \cite{zhu2018haze}                 & 21.40          & 0.923          & 0.090             & 17.93          & 0.891          & 0.132             \\
			DehazeNet \cite{cai2016dehazenet}              & 15.08          & 0.778          & 0.134             & 15.66          & 0.810          & 0.192             \\
			MSCNN \cite{ren2020single}                  & 17.83          & 0.855          & 0.092             & 20.53          & 0.926          & 0.076             \\
			AODNet \cite{li2017aod}                 & 18.05          & 0.860          & 0.093             & 18.49          & 0.863          & 0.152             \\
			GCANet \cite{chen2019gated}                 & 20.98          & 0.906          & 0.180             & 16.39          & 0.831          & 0.215             \\
			FFANet \cite{qin2020ffa}                 & 24.13          & 0.938          & 0.038             & 18.71          & 0.882          & 0.126             \\
			DADFNet                 & \textbf{24.56}          & \textbf{0.951}          & \textbf{0.033}             & \textbf{24.19}          & \textbf{0.948}          & \textbf{0.053}            \\ \hline
		\end{tabular} \label{table:metric}
	\end{table*}
	\subsection{Experimental Settings}
	%
	%
	%
	%
	%
	%
	%
	In the proposed DADFNet, five trade-off parameters of loss function are empirically set as $\lambda_1 = 1$, $\lambda_2 = 0.5$, $\lambda_3 = 0.5$, $\lambda_4 = 0.01$, and $\lambda_5 = 0.0005$. We set total epoch and batch size to $200$ and $4$, respectively. Meanwhile, we adopt Adam technique for optimization with initial learning rate being $10^{-4}$. In every $30$ epoch, the learning rate is reduced to $1/2$ of current learning rate. For better application in intelligent transportation system, we select $2000$ images from RESIDE \cite{li2018benchmarking} and $2000$ images from the Seaships \cite{shao2018seaships}. In the training, each image in the train dataset is divided into several $256 \times 256$ image patches. In each epoch, we adopt the atmospheric scattering model to synthesize haze. Let $J$ and $d$ be the clear image and depth information, respectively, the hazy image $I$ can be generated by
	\begin{equation}\label{Eq:ASMl}
	I = e^{-\beta d}(J - 1) + A,
	\end{equation}
	where $\beta$ and $A$ denote the scattering coefficient and atmospheric light value, respectively. Since the Seaships dataset lacks depth information, we adopt Mega-Depth \cite{li2018megadepth} to produce scene depth. Moreover, we synthesize hazy image of different degrees by setting $\beta \in [0.08, 0.3]$ and $A \in [0.7, 1.0]$. In particular, we implement the proposed DADFNet by the PyTorch platform of Python $3.7$. All experiments are running on a PC with Intel (R) Core (TM) i$5$-$10600$KF CPU @ $4.10$GHz and Nvidia GeForce GTX $2080$ Ti GPU.
	\def \fc{1.66cm} 
	\def \he{-0.1cm}
	\begin{figure*}[h]
		\centering
		
		\setlength{\abovecaptionskip}{0.cm}
		\subfigure[]{\includegraphics[width=\fc]{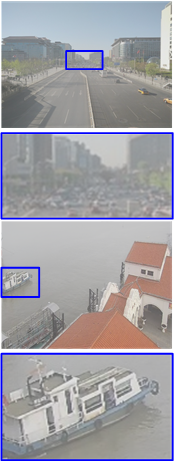}}
		\subfigure[]{\includegraphics[width=\fc]{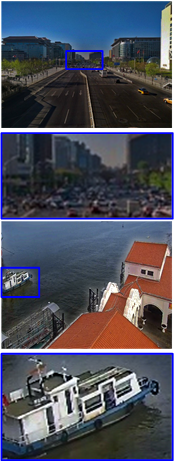}}
		\subfigure[]{\includegraphics[width=\fc]{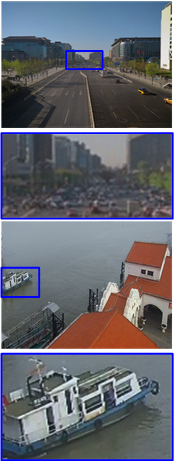}}
		\subfigure[]{\includegraphics[width=\fc]{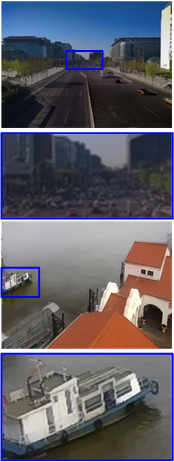}}
		\subfigure[]{\includegraphics[width=\fc]{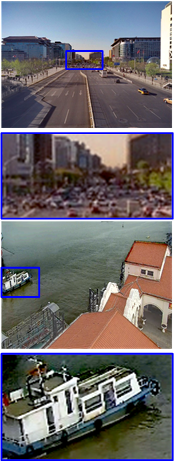}}
		\subfigure[]{\includegraphics[width=\fc]{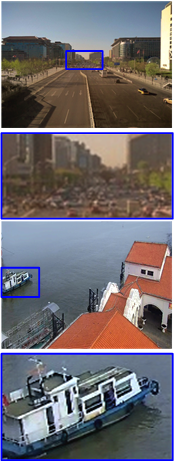}}
		\subfigure[]{\includegraphics[width=\fc]{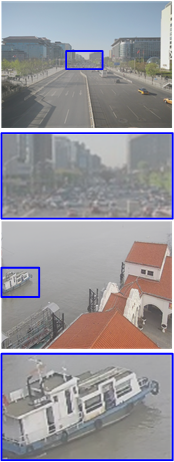}}\\
		\subfigure[]{\includegraphics[width=\fc]{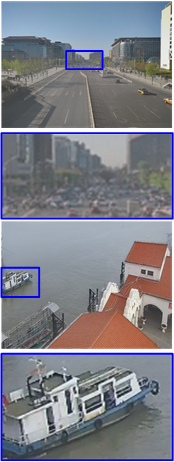}}
		\subfigure[]{\includegraphics[width=\fc]{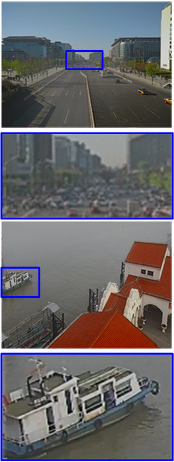}}
		\subfigure[]{\includegraphics[width=\fc]{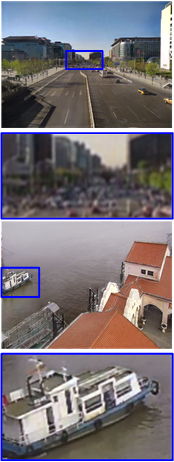}}
		\subfigure[]{\includegraphics[width=\fc]{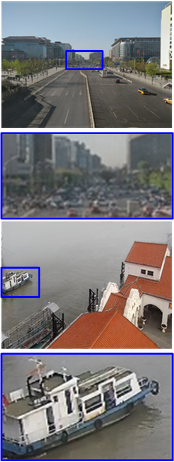}}
		\subfigure[]{\includegraphics[width=\fc]{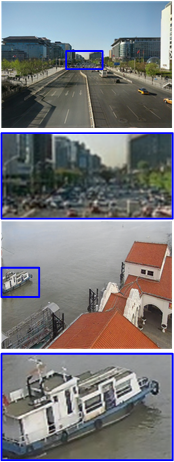}}
		\subfigure[]{\includegraphics[width=\fc]{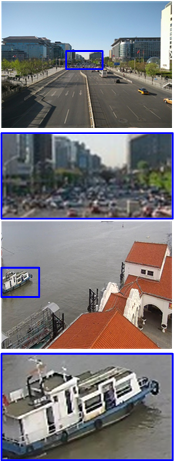}}
		
		\caption{Visual comparison of various methods on the RESIDE and Seaships. (a) Hazy inputs. (b) DCP \cite{he2010single}. (c) CAP \cite{zhu2015fast}. (d) GRM \cite{chen2016robust}. (e) HL \cite{berman2018single}. (f) F-LDCP \cite{zhu2018haze}. (g) DehazeNet \cite{cai2016dehazenet}. (h) MSCNN \cite{ren2020single}. (i) AODNet \cite{li2017aod}. (j) GCANet \cite{chen2019gated}. (k) FFANet \cite{qin2020ffa}. (l) DADFNet. (m) Ground truths.}
		\label{fig:syn}
	\end{figure*}
	\def \fc{1.9cm} 
	\def \he{-0.1cm}
	\begin{figure*}[t]
		\centering
		
		\subfigure[]{\includegraphics[width=\fc]{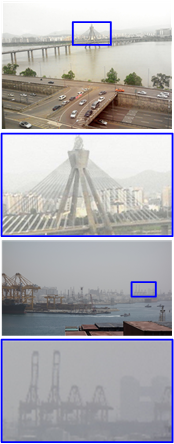}}
		\subfigure[]{\includegraphics[width=\fc]{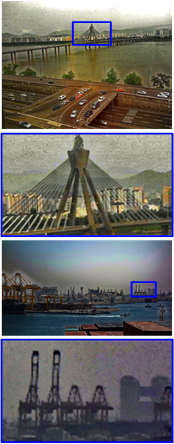}}
		\subfigure[]{\includegraphics[width=\fc]{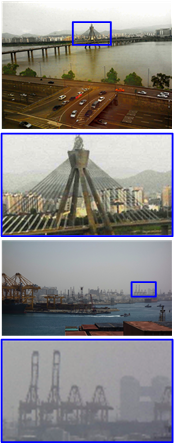}}
		\subfigure[]{\includegraphics[width=\fc]{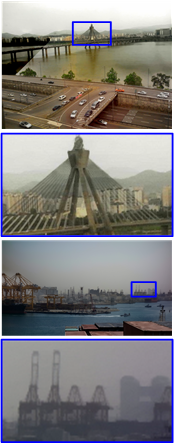}}
		\subfigure[]{\includegraphics[width=\fc]{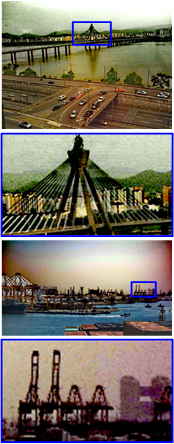}}
		\subfigure[]{\includegraphics[width=\fc]{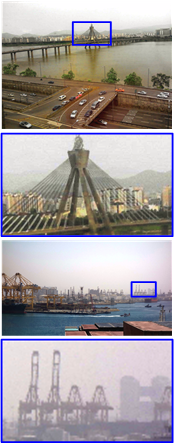}}\\
		\subfigure[]{\includegraphics[width=\fc]{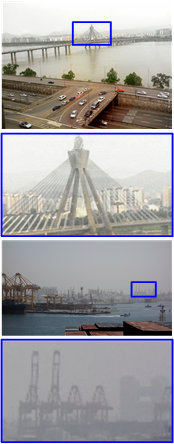}}
		\subfigure[]{\includegraphics[width=\fc]{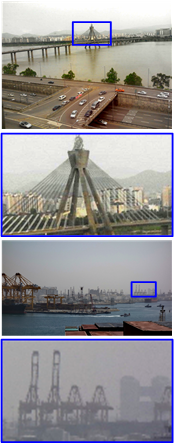}}
		\subfigure[]{\includegraphics[width=\fc]{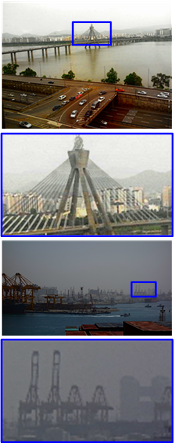}}
		\subfigure[]{\includegraphics[width=\fc]{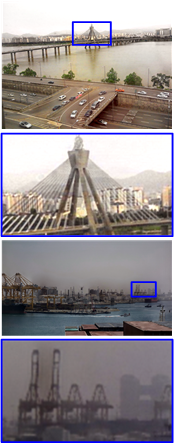}}
		\subfigure[]{\includegraphics[width=\fc]{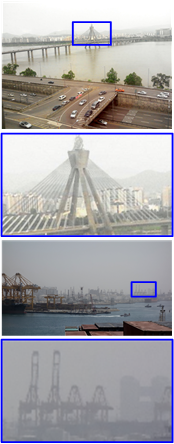}}
		\subfigure[]{\includegraphics[width=\fc]{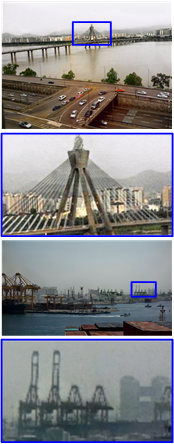}}
		
		\caption{Visual comparison of various methods on the real-world hazy images. (a) Hazy inputs. (b) DCP \cite{he2010single}. (c) CAP \cite{zhu2015fast}. (d) GRM \cite{chen2016robust}. (e) HL \cite{berman2018single}. (f) F-LDCP \cite{zhu2018haze}. (g) DehazeNet \cite{cai2016dehazenet}. (h) MSCNN \cite{ren2020single}. (i) AODNet \cite{li2017aod}. (j) GCANet \cite{chen2019gated}. (k) FFANet \cite{qin2020ffa}. (l) DADFNet.}
		\label{fig:real}
	\end{figure*}
	\subsection{Quantitative Results on Benchmarks}
	In this section, we compare our DADFNet with several state-of-the-art methods on $50$ RESIDE and $50$ Seaships images. The hazy images are synthesized by Eq. (\ref{Eq:ASMl}) with $\beta \in [0.08, 0.3]$ and $A \in [0.7, 1.0]$.  Meanwhile, we adopt three metrics (i.e., PSNR, SSIM, and LPIPS) to compare the image reconstruction performance of all methods. 
	
	Table \ref{table:metric} reports the evaluation results of all methods on RESIDE and Seaships. In terms of RESIDE, our method takes first place with $24.56$ dB PSNR, $0.951$ SSIM, and $0.033$ LPIPS, which is superior to FFANet trained on the RESIDE. Meanwhile, we implement comparison of various methods on synthetic maritime hazy images. In this comparison, our DADFNet also gains the best performance, which achieves the $3.11$ dB PSNR, $0.019$ SSIM, and $0.017$ LPIPS improvement compared to the suboptimal method (CAP). Meanwhile, Fig. \ref{fig:syn} shows the imaging quality of restored images on both RESIDE and Seaships. We can find that CAP, DehazeNet, AODNet, and FFANet fail to suppress the dense haze, making the vital object invisible. Although DCP and GRM have satisfactory image dehazing performance, these methods tend to reconstruct the low-illumination scenes for degrading the image further. Intuitively, the enhanced images of HL, F-LDCP, MSCNN, and GCANet are able to produce haze-free scenarios. However, the visual quality of these results will be affected by color distortion and artifacts. Owing to the strong feature extraction ability, our DADFNet can fully eliminate the haze and persevere the scene naturalness.
	\begin{figure*}[t]
		\centering
		\includegraphics[width=1\linewidth]{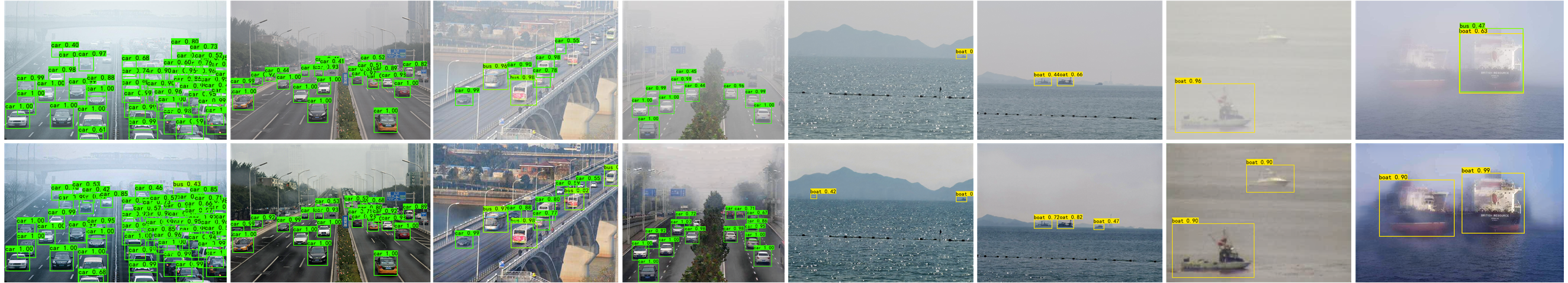}
		\caption{The detection results on real-world hazy images  (Top) and restored images yielded by our DADFNet (Bottom).}
		\label{fig:detectionexp}
	\end{figure*}
	\begin{table}[t]
		\setlength{\tabcolsep}{5pt}
		\centering
		\small
		\caption{Detection average precision (\%) for different dehazing methods on $276$ real-world hazy images. The best results are in \textbf{bold}, and the second best are with \underline{underline}.}
		\begin{tabular}{ccccccc|c}
			\hline
			Method                           & Bicycle & Boat  & Bus   & Car   & Motorbike & Person & mAP$\uparrow$ \\ \hline \hline
			Hazy                             & 63.77   & 76.45 & 43.75 & 71.81 & 42.37     & 61.90  & 60.01         \\
			DCP\cite{he2010single}           & 67.44   & \textbf{87.45} & \underline{50.70} & 73.77 & 41.47     & 62.36  & \underline{63.86}         \\
			CAP\cite{zhu2015fast}            & 57.79   & 81.60 & 46.97 & 72.42 & 35.73     & 58.35  & 58.81         \\
			GRM\cite{chen2016robust}         & 25.65   & 71.37 & 40.23 & 62.02 & 27.87     & 46.48  & 45.60         \\
			HL\cite{berman2018single}        & 64.41   & 82.97 & 49.50 & \underline{74.15} & 40.18     & 60.80  & 62.00         \\
			F-LDCP\cite{zhu2018haze}         & 63.68   & 85.94 & \textbf{50.83} & 73.66 & 41.81     & \textbf{63.64}  & 63.26         \\
			DehazeNet\cite{cai2016dehazenet} & \underline{72.35}   & 78.75 & 44.99 & 72.74 & \underline{43.67}     & \underline{62.91}  & 62.57         \\
			MSCNN\cite{ren2020single}        & 71.25   & 84.42 & 50.63 & 73.93 & 40.04     & 62.57  & 63.81         \\
			AODNet\cite{li2017aod}           & 66.67   & \underline{86.64} & 47.32 & 74.07 & 40.56     & 61.29  & 62.76         \\
			GCANet\cite{chen2019gated}       & 63.69   & 83.92 & 48.33 & 73.13 & 36.11     & 60.93  & 61.02         \\
			FFANet\cite{qin2020ffa}          & 71.87   & 76.78 & 45.13 & 73.05 & \textbf{44.31}     & 62.10  & 62.21         \\
			DADFNet                          & \textbf{78.36}   & 85.22 & 49.47 & \textbf{74.86} & 42.98     & 62.28  & \textbf{65.53}   \\ \hline     
		\end{tabular}\label{table:mAP}
	\end{table}
	\subsection{Qualitative Results on Real-World Images}
	To verify that our method can enhance the visibility of important information in ITS environments, we conduct a dehazing experiment on outdoor traffic scenes shown in Fig. \ref{fig:real}. The haze-free versions obtained by DCP and HL suffer from severe color distortion. Obviously, DCP, HL, AODNet, and GCANet easily cause critical information hidden in the dark, which takes the challenge for semantic analysis and target observation. Furthermore, CAP, GRM, F-LDCP, DehazeNet, MSCNN, and FFANet fail to robustly handle complex haze in the real world, resulting in the hazy residue. Compared with other methods, our proposed dehazer has superior imaging performance. The robust haze suppression capability benefits from the precise hazy distribution generation of dual attention and the strong feature extraction of the high-low frequent-guided network.
	
	Meanwhile, we analyze the influence of various dehazers on target detection in ITS. We select $276$ real-world hazy images as the test dataset. Especially, hazy and enhanced images are fed into the YOLOv4 trained on the VOC dataset. Table \ref{table:mAP} displays the mAP results on the real-world hazy images and the images enhanced by various methods. We can find that DADFNet has the best mAP, higher than the sub-optimal method (DCP) by $1.67\%$. In particular, our DADFNet has the optimal detection performance on the ``Bicycle'' and ``Car'', which are the essential objects in ITS. In the detection experiments of other objects, our method only fails behind the optimal result with a small difference. Furthermore, Fig. \ref{fig:detectionexp} shows the detection versions of the hazy images and the images enhanced by our DADFNet on the road and maritime scenes. Owing to the production of haze, the imaging scene becomes unclear and ambiguous. This degradation seriously destroys the key information of objects, bringing critical challenges for target detection. Therefore, target detection on the hazy images easily suffers from misdetection, especially for small-scale targets. In contrast, our proposed hazy elimination strategy can solve the above issues to improve the detection effect.
	
	It is worth mentioning that we conduct a running time analysis. Our DADFNet processes a $1,920 \times 1,080$ image requiring only $6.3$ ms in RTX $2080$ Ti GPU and $9.6$ ms in GTX $1050$ GPU. Therefore, our proposed method can achieve real-time dehazing on several computing devices to satisfy different task requirements.
	\section{Conclusion}
	To conclude, we have proposed a dual attention and dual frequency-guided dehazing network (DADFNet) to implement efficient and effective visibility enhancement. It accordingly contributes to flexible transportation surveillance under hazy weathers. The major contributions of this work were threefold. First, an efficient dehazer, mainly including DAM and HLFN modules, was presented to real-timely perform visibility enhancement in video-empowered ITSs. Second, the proposed DAM module is capable of accurately modeling the hazy distribution, leading to elimination of haze effects. Third, the HLFN is able to jointly preserve the low-frequency structures and high-frequency details in restored images. Comprehensive experiments have demonstrated the superior imaging performance of our DADFNet under hazy atmosphere conditions.
	%
	%
	%
	%
	\bibliographystyle{splncs04}
	\bibliography{aaai22.bib}
	
\end{document}